\documentclass{article}

\PassOptionsToPackage{}{natbib}
\usepackage[final]{icbinb2021}

\usepackage[utf8]{inputenc} 
\usepackage[T1]{fontenc}    
\usepackage{hyperref}       
\usepackage{url}            
\usepackage{booktabs}       
\usepackage{amsfonts}       
\usepackage{nicefrac}       
\usepackage{microtype}      
\usepackage{xcolor}         
\usepackage{algorithm}
\usepackage{algorithmic}
\usepackage{multirow}
\usepackage{multicol}
\usepackage{float}

\title{Forecasting Market Prices using DL with Data Augmentation and Meta-learning: ARIMA still wins!}

\author{%
  Vedant Shah \\
  APPCAIR \\
  BITS Pilani, K K Birla Goa Campus \\
  \texttt{f20180566@goa.bits-pilani.ac.in} \\
   \And
   Gautam Shroff \\
   TCS Research \\
   New Delhi \\
   \texttt{gautam.shroff@tcs.com} \\
}

\begin{document}

\maketitle

\begin{abstract}
Deep-learning techniques have been successfully used for time-series forecasting and have often shown superior performance
on many standard benchmark datasets as compared to traditional techniques. Here we present a comprehensive and comparative 
study of performance of deep-learning techniques for forecasting prices in financial markets. We benchmark state-of-the-art
deep-learning baselines, such as NBeats, etc., on data from currency as well as stock markets. We also generate synthetic data using a fuzzy-logic
based model of demand driven by technical rules such as moving averages, which are often used by traders. We benchmark
the baseline techniques on this synthetic data as well as use it for data augmentation. We also apply gradient-based 
meta-learning to account for non-stationarity of financial time-series. Our extensive experiments notwithstanding, the
surprising result is that the standard ARIMA models outperforms deep-learning even using data augmentation or meta-learning. We
conclude by speculating as to why this might be the case.

\end{abstract}

\section{Introduction}

Forecasting prices in financial markets has long been attempted using
a variety of statistical as well as modern machine-learning techniques;
in theory markets are `efficient', and should not be predictable, making
this task potentially ill-posed to start with. Nevertheless, financial
markets are rife with algorithmic trading engines that aim at precisely this, attempting to discover any lingering inefficiencies that could 
yield a usable trading signal.

The behaviour of stock markets is a result of complex combination of several trading heuristics followed by a large number of groups of traders ranging from big financial firms to individual investors, each with varying impact. Previous works (\cite{6824795, 6873314}) have shown that these trading strategies and the behaviour arising from their combination can be modeled as a set of mathematical equations for the excess demand function using fuzzy logic. These set of equations are capable of modelling the real-world financial markets to a good extent as shown in \cite{6965624}. These models can then be used to generate data of desired amount which will follow the behaviour arising from the combination of all the trading heuristics considered while building the model.\par

Present day deep learning models require large amounts of data to train. The foremost problem in using deep-learning for market data forecasting is the lack of availability of large datasets; after all, there is only
so much historical data available. We hypothesize that synthetic data can be used to improve the performance of existing Deep Learning models on forecasting real-world market data. This is based on the premise that training on even very dissimilar data can help in improving the performance of deep learning models as shown in \cite{Malhotra2017TimeNetPD}.\par

Meta-learning is a machine learning paradigm wherein a model is trained on multiple similar tasks; for example, tasks having slightly different data distributions. Gradient-based meta-learning using MAML introduced in \cite{Finn2017ModelAgnosticMF} is a model-agnostic procedure wherein a base model is adapted using the training
data of each task via a few gradient steps. During meta-training, the based model is trained in an outer-loop, again using gradient descent, so that its post-adaptation performance on the test data of the meta-training tasks is optimized.
The goal is to initialize the model with a set of parameters which can easily be adapted to different downstream tasks, potentially helping the model to generalize to out of distribution data. Financial data can be thought of as undergoing frequent changes of distribution as economic conditions and
investor sentiments fluctuate due to global and national events\par

We present a comprehensive study in which we attempt using meta-learning and synthetically generated data for data augmentation,
to improve performance of deep-learning models for forecasting financial market data. We report results across various deep learning models: MLPs, LSTMs, Transformers (\cite{Vaswani2017AttentionIA}) and two state of the art time-series forecasting models: NBEATS (\cite{Oreshkin2020NBEATSNB}) and Temporal Fusion Transformers (TFT) (\cite{Lim2019TemporalFT}). Surprisingly,
we find that none of these models is better than the traditional ARIMA model, even using data augmentation or applying meta-learning.\par

\section{Related work}
\cite{MAKRIDAKIS2000451} and \cite{Makridakis2018StatisticalAM} provide an in depth review of the performance of various statistical and Machine Learning algorithms for time-series forecasting in the M3 competition. Recently, \cite{Oreshkin2020NBEATSNB} proposed N-BEATS, a deep learning univariate time-series forecasting model which is capable of beating statistical approaches on the M4 benchmark. \cite{Lim2019TemporalFT} introduced Temporal Fusion Transformer, a state of the art multivariate time-series forecasting model which shows improved performance as compared to ARIMA and ETS on a number of real-world tasks. DeepAR proposed in \cite{Flunkert2017DeepARPF} uses auto-regressive RNN for probabilistic forecasting. \cite{Oreshkin2021MetalearningFW} studies zero-shot learning for univariate time series forecasting using Model Agnostic Meta Learning as proposed in \cite{Finn2017ModelAgnosticMF} on N-BEATS. Another very recent work, \cite{Elsayed2021DoWR} shows that traditional forecasting approaches such as Gradient Tree Boosting Models are still capable of beating a number of state of the art deep learning approached on an array of different datasets. 

\section{Methodology}
\label{method}

\cite{6824795} uses fuzzy logic to develop a mathematical model of the excess demand of the stock market by taking different trading heuristics into account. This mathematical is then used to generate synthetic data using:
 \begin{equation}
 \label{eq1}
    \ln(p_{t+1}) = \ln(p_{t}) + \sum_{i=0}^{M}a_{i}(t)ed_{i}(x_{t}) 
 \end{equation}

where $p_t$ denotes the price of a share at a give time-step, $a_{i}(t)$ denotes the influence of a particular heuristic, and $ed_{i}(x_{t})$ denotes the excess-demand function for the heuristic with $x_{t}$ being a time dependent variable. We consider synthetic data generated from \textbf{Heuristic 1} discussed in \cite{6824795} for all purposes. Using fuzzy logic, excess demand can be modeled into the expression given below; twelve such heuristics
are discussed in \cite{6824795}.

\begin{equation}
\label{eq2}
    ed_{1}(x_{1, t}^{(m, n)}) = \frac{\Sigma_{i=1}^{7}c_{i}\mu_{A_{i}}(x_{1, t}^{(m, n)})}{\Sigma_{i=1}^{7}\mu_{A_{i}}(x_{1, t}^{(m, n)})}
\end{equation}

where $x_{1, t}^{(m, n)}$ is the ratio of the natural logarithm of the two moving averages of length $m$ and $n$ of the stock price. The price of a stock at time step $t$ is denoted by $p_t$. $\mu_{A_{i}}$ are a fuzzy membership functions for each of the 7 rules as in \cite{6824795}, describing
when each rule applies. The moving average of length $n$ and the ratio of the natural logarithms of two moving averages of length $m$ and $n$ can then be calculated as: 
$\bar{p}_{t, n} = \frac{1}{n}\sum_{i=0}^{n-1}p_{t-1}$
and $
    x_{1, t}^{(m, n)} = \ln(\frac{\bar{p}_{t,m}}{\bar{p}_{t,n}})
$

We experiment with two different techniques in an attempt to improve market data forecasting using synthetic data:\par

\textbf{Data Augmentation:} We augment the real world data with an equal amount of synthetic data during training the models. For each epoch, batches of real-world data and synthetic data are fed randomly to the deep learning model until all the batches of both synthetic and real world data are exhausted. We use a lookback length of 20 for the real-world data and a lookback length of 5 for the synthetic data. For models with fixed input length such as MLP and NBEATS, we interpolate the lookback windows of the synthetic data to make their length equal to those of the real-world data. The algorithm description can be found in the appendix.\par

\textbf{Meta Learning:} We create the tasks for meta-training from the training part of a dataset and those for meta-testing from testing part of the dataset. For synthetic data, these divisions are made within each of the 40 time-series. For creating each task for meta-training, a time-step $d$ is chosen randomly from the training data. Let $l$ be the total length of the window ($lookback + prediction$). For each such time-step $d$, $k$ such windows, $[d-l-j, d-j]$ are sampled randomly from before $d$, where $j \sim U[1, a]$. This forms the training set of the task. Similarly, $r$ windows are sampled randomly after $d$ which forms the test set of the task. The same procedure is used on the testing dataset to create meta-testing tasks. For synthetic data, meta-training and meta-testing tasks are created from each series and then merged. All the deep learning models are then trained using meta-learning on both real-world and synthetic data individually using First Order Model Agnostic Meta Learning algorithm (\cite{Nichol2018OnFM}).

\section{Empirical evaluation}
\label{evaluation}

\subsection{Materials}
\label{dataset}
We use run experiments on three datasets: one \textbf{synthetic} and two \textbf{real world} datasets.\par

\textbf{Synthetic Data:} We use equations (\ref{eq1}) and (\ref{eq2}) with $(m, n) = (1, 5)$ to generate 40 synthetic time series each containing 500 time-steps. Each of these time-series is initialized with different seeds and a random-walk for 100 time-steps. For normal training and testing, 36 time-series are used as training data and 2 time-series are used for validation and testing data each. For meta learning, the first 400 time-steps of each time series are used as the meta-training data and the last 100 time-steps are used as the meta-testing data.

\textbf{Banknifty Data:} Banknifty data is a time-series of closing values of the banknifty index, recorded every week day from 1st January, 2016 to 31st May, 2021. After cleaning the data we split the data into a training time series containing 1050 time-steps and testing and validation time series, each containing 144 time-steps. For meta learning the the time series consisting of 1050 time-steps is used to create tasks for meta-training and the rest of the time series consisting of 288 time-steps (144 + 144) is used for creating meta-testing tests.

\textbf{Forex Data:} Forex Data is a time-series of spot prices of 1 INR in USD, recorded every weekday from 1st January, 2010 to 14th August, 2020. The total length of the time series is 2562 time-steps of which the the first 2050 time-steps are used as the training data, time steps 2045 to 2306 are used as the validation data and time-steps 2301 to 2562 are used as testing data. For meta learning, the first 2050 time-steps are used for creating meta-training tasks and the remaining data is used to create meta-testing tasks.

\subsection{Results}
\label{results}

\begin{table}
  \caption{Results on three datasets \textbf{SN} - Synthetic, \textbf{FR} - Forex, and \textbf{BN} - Banknifty for \textbf{NTR} - Normal Training, \textbf{DA} - Data Augmented Training and \textbf{M2L} - Meta Learning. \textbf{OS} - One step prediction, \textbf{TS} - Tens step prediction, \textbf{Average DL} - Average of performance of all five deep learning models}
  \label{table1}
  \centering
  \begin{tabular}{llllllllllll}
    \toprule
                            &               &           &           &            & \multicolumn{2}{c}{ARIMA} & \multicolumn{2}{c}{Average DL} & \multicolumn{2}{c}{Best DL Model} &  Best DL \\
    \cmidrule(r){6-11}
                            &               &           &           &            &  RMSE  &  MAPE  &  RMSE  &  MAPE  &  RMSE  &  MAPE  &                                       Model Name \\
    \cmidrule(r){1-12}
    \multirow{4}{*}{SN}     &  \multicolumn{2}{c}{NTR}  & \multicolumn{2}{c}{OS} & 0.5496 & 0.0387 & 0.3837 & 0.0127 & \textbf{0.2313} & \textbf{0.0102} & NBEATS      \\  
                            &               &           & \multicolumn{2}{c}{TS} & 0.8561 & 0.0527 & 0.8126 & 0.0454 & \textbf{0.7591} & \textbf{0.0409} & NBEATS      \\
    \cmidrule(r){2-12}
                            &  \multicolumn{2}{c}{M2L}  & \multicolumn{2}{c}{OS} &   -    &    -   & 0.3655 & 0.0262 & 0.2961 & 0.0219 & NBEATS                         \\
                            &               &           & \multicolumn{2}{c}{TS} &   -    &    -   & 0.7510 & 0.0530 & 0.7205 & 0.0514 & LSTM                           \\
    \midrule
    \multirow{6}{*}{FR}     &  \multicolumn{2}{c}{NTR}  & \multicolumn{2}{c}{OS} & \textbf{0.1982} & \textbf{0.0027} & 0.2962 & 0.0029 & 0.2734 & 0.0027 & LSTM        \\  
                            &               &           & \multicolumn{2}{c}{TS} & \textbf{0.4754} & \textbf{0.0056} & 0.5923 & 0.0058 & 0.5648 & 0.0055 & LSTM        \\
    \cmidrule(r){2-12}
                            &  \multicolumn{2}{c}{DA}   & \multicolumn{2}{c}{OS} &   -    &    -   & 0.2849 & 0.0028 & 0.2609 & 0.0027 & Transformer \\  
                            &               &           & \multicolumn{2}{c}{TS} &   -    &    -   & 0.5980 & 0.0059 & 0.5680 & 0.0055 &  LSTM       \\                        
    \cmidrule(r){2-12}
                            &  \multicolumn{2}{c}{M2L}  & \multicolumn{2}{c}{OS} &   -    &    -   & 0.3443 & 0.0038 & 0.3235 & 0.0036 &  MLP        \\
                            &               &           & \multicolumn{2}{c}{TS} &   -    &    -   & 0.6579 & 0.0069 & 0.6029 & 0.0064 &  LSTM       \\
    \midrule
    \multirow{6}{*}{BN}     &  \multicolumn{2}{c}{NTR}  & \multicolumn{2}{c}{OS} & \textbf{435.18} & \textbf{0.0131}  & 816.63  & 0.0187 & 704.68  & 0.0158 & NBEATS     \\  
                            &               &           & \multicolumn{2}{c}{TS} & \textbf{1196.09} & \textbf{0.0314} & 1491.46 & 0.0337 & 1396.34 & 0.0312 & LSTM      \\
    \cmidrule(r){2-12}
                            &  \multicolumn{2}{c}{DA}   & \multicolumn{2}{c}{OS} &    -   &    -   & 858.85 & 0.0200 & 728.65 & 0.0165 &  MLP        \\  
                            &               &           & \multicolumn{2}{c}{TS} &    -   &    -   & 1486.22 & 0.0339 & 1397.29 & 0.0313 & LSTM       \\                        
    \cmidrule(r){2-12}
                            &  \multicolumn{2}{c}{M2L}  & \multicolumn{2}{c}{OS} &    -   &    -   & 877.94 & 0.0238 & 784.62 & 0.0213 & MLP         \\
                            &               &           & \multicolumn{2}{c}{TS} &    -   &    -   & 1674.17 & 0.0434 & 1562.72 & 0.0403 & LSTM       \\
    \bottomrule
  \end{tabular}
\end{table}

Table \ref{table1} compares\footnote{Transformer suffers from exploding gradients when Meta Learning. TFT suffers from exploding gradients when Meta Learning on Banknifty Data. These cases aren't considered while calculating average performances.} the performance of deep learning models trained using different training techniques with ARIMA. We use a multitude of deep learning models ranging from simple models to current state of the art models for time-series forecasting. Results were obtained for: MLP, LSTM, Transformers , NBEATS and TFT. All the experiments were performed using a 12GB NVIDIA Tesla K80 GPU, 12GB RAM and an Intel(R) Xeon(R) CPU with 2 cores. The results have been reported for two well known metrics: \textbf{Root Mean Squared Error (RMSE)} and \textbf{Mean Absolute Percentage Error (MAPE)}.

The lookback window length for the synthetic dataset has been taken as 5 since each time-step depends on the last 5 time-steps and the lookback window length for both the real-world datasets has been taken as 20 which was arrived at by experimenting with different lookback lengths. The lookback windows are locally normalized before passing through the model which allows the real-world and the synthetic data to be combined during data augmentation. We evaluate each model for a prediction horizon of 10 time-steps and the results are calculated after de-normalizing the predictions. We further decompose the results obtained to get results for one-step ahead prediction, five step ahead prediction and 6th to 10th step prediction. Results for one step (OS) and ten step  (TS) ahead predictions have been reported.\par

\subsubsection{Rollout Testing}
We further experiment with try rollout testing for models trained with and without data augmentation on non-autoregressive models (MLP, NBEATS and TFT). In rollout testing, we take into account the previous $n$ points to predict the next time-step where $n$ is the length of the lookback window. The lookback window is of fixed length and keeps sliding one step at a time as and when new predictions are made. We take $n$ = 5 for synthetic data in an attempt to make it easier for the predictor to capture the actual relationship between the time-steps (remember that we use $n$ = 5 in equation \ref{eq2} for generating the data). $n$ is taken as 20 for Banknifty and Forex datasets. The results are presented in Table \ref{table2}.\par

\begin{table}
  \caption{Rollout Testing with and without Data Augmentation for non-autoregressive models}
  \label{table2}
  \centering
  \begin{tabular}{lllllllllll}
    \toprule
                            &               &           &           &            & \multicolumn{2}{c}{MLP} & \multicolumn{2}{c}{NBEATS} & \multicolumn{2}{c}{TFT}  \\
    \cmidrule(r){6-11}
                            &               &           &           &            &  RMSE  &  MAPE  &  RMSE  &  MAPE  &  RMSE  &  MAPE                              \\
    \midrule
    \multirow{2}{*}{SN}     &               &           & \multicolumn{2}{c}{OS} & 0.2754 & 0.0127 & 0.2315 & 0.0102 & 0.4361 & 0.0225                             \\  
                            &               &           & \multicolumn{2}{c}{TS} & 0.9691 & 0.0522 & 0.9957 & 0.0529 & 0.8567 & 0.0479                             \\
    \midrule
    \multirow{4}{*}{FR}     &  \multicolumn{2}{c}{NTR}  & \multicolumn{2}{c}{OS} & 0.2935 & 0.0029 & 0.3233 & 0.0033 & 0.3936 & 0.0038                          \\  
                            &               &              & \multicolumn{2}{c}{TS} & 0.6030 & 0.0059 & 0.7699 & 0.0079 & 0.7036 & 0.0068                          \\
    \cmidrule(r){2-11}
                            &  \multicolumn{2}{c}{DA}   & \multicolumn{2}{c}{OS} & 0.3093 & 0.0030 & 0.2984 & 0.0029 & 0.3005 & 0.0030                          \\  
                            &               &           & \multicolumn{2}{c}{TS} & 0.6877 & 0.0064 & 0.5982 & 0.0058 & 0.6784 & 0.0067                             \\                        
    \midrule
    \multirow{4}{*}{BN}     &  \multicolumn{2}{c}{NTR}  & \multicolumn{2}{c}{OS} & 730.23 & 0.0164 & 707.82 & 0.0159 & 977.06 & 0.0233                          \\  
                            &               &           & \multicolumn{2}{c}{TS} & 1480.58 & 0.0309 & 1413.56 & 0.0318 & 1431.35 & 0.0339                          \\
    \cmidrule(r){2-11}
                            &  \multicolumn{2}{c}{DA}   & \multicolumn{2}{c}{OS} & 727.78 & 0.0165 & 851.85 & 0.0209 & 1006.77 & 0.0239                        \\  
                            &               &           & \multicolumn{2}{c}{TS} & 1462.83 & 0.0306 & 1551.39 & 0.0374 & 1583.22 & 0.0371                          \\                        
    \bottomrule
  \end{tabular}
\end{table}

 Clearly, state of the art deep learning models for time-series forecasting fail to beat classical approaches such as ARIMA on the real-world data, even after data augmentation and meta learning. Deep Learning models successfully beat ARIMA on synthetic data which is deterministic. Real-world financial data is very noisy and chaotic. Machine Learning models rely on a global feedback signal or a recent signal for optimization. Due to the (financial) data being noisy and chaotic and the fact that deep learning models are often trained on multiple time-series, the feedback signals are also very noisy leading to poor optimization. The synthetic data on the other hand is deterministic, hence leading to noiseless signals due to which deep learning models perform better on synthetic data. ARIMA on the other hand relies on a purely local signal computed every time it’s applied. A local signal is much stronger when there is more noise in the longer signals, e.g., because the
inefficiencies that lead to potential trading signals are \textit{not}
similar over time and change continuously.

\section{Conclusion}

We present a comprehensive and comparative study on performance of different Deep Learning techniques and ARIMA, a classical approach on market data forecasting. While current state of the art deep models are
often capable of outperforming classical forecasting techniques, they fail to do so on financial data - an initally surprising negative result. We attribute the reason to the fact that financial data is inherently chaotic. Autoregressive models such as ARIMA in general perform better for prediction when there is less reason to believe that the regularities enabling prediction are themselves non-stationary and cannot therefore be exploited by machine-learning models. We find that this is indeed true for the financial data that we experimented on.

\bibliographystyle{apalike}
\bibliography{references}


\clearpage
\appendix

\section{Appendix}

\subsection{Data Augmentation}

\begin{algorithm}
\caption{Data Augmentation}
\begin{algorithmic}[1]
\STATE Initialize the synthetic dataloader $\mathcal{S}$, the real-world dataloader $\mathcal{R}$ with appropriate batch sizes and the number of epochs $e$
\STATE $n \leftarrow len(\mathcal{S})$
\STATE $m \leftarrow len(\mathcal{R})$
\FOR{$i$ in $e$}
\STATE $s \leftarrow 0$
\STATE $r \leftarrow 0$
\WHILE{$s < n$ or $r < m$}
\STATE $a \sim \mathit{U}(0, 1)$
\IF{$a > 0.5$}
\IF{$s < n$}
\STATE $batch \sim \mathcal{S}$
\STATE Input the lookback window into the model
\STATE Calculate the loss on the batch
\STATE Backpropagate on the loss
\STATE $s \leftarrow s+1$
\ENDIF
\ELSE
\IF{$r < m$}
\STATE $batch \sim \mathcal{R}$
\STATE Input the lookback window into the model
\STATE Calculate the loss on the batch
\STATE Backpropagate on the loss
\STATE $r \leftarrow r+1$
\ENDIF
\ENDIF
\ENDWHILE
\ENDFOR
\end{algorithmic}
\end{algorithm}

\subsection{Model Configurations and Hyperparameters}
We use \hyperref{https://pytorch-forecasting.readthedocs.io/en/latest/}{PyTorch Forecasting} for MLP, LSTM, NBEATS and TFT and data loading and pre-processing utilities. For Transformer, we use the PyTorch's nn.Transformer. Any parameters not included here have been used with their default values as defined in the respective libraries.

We use 5 fast adaptation steps for MAML throughout all the experiments. 

\subsubsection{Transformer}
\begin{table}[H]
    \centering
    \begin{tabular}{c|c}
        \toprule
        learning rate & 1e-4  \\
        batch size & 64 \\
        weight decay & 1e-2 \\
        $d_{model}$ & 256 \\
        attn heads & 8 \\
        encoders & 4 \\
        decoders & 4 \\
        feedforward dim & 512 \\
        MAML lr & 1e-1 \\
        learner lr (SN, BN, FR) & 1e-5, 1e-3, 1e-4 \\
        \bottomrule
    \end{tabular}
\end{table}

\subsubsection{NBEATS}
\begin{table}[H]
    \centering
    \begin{tabular}{c|c}
        \toprule
        learning rate & 1e-4  \\
        batch size & 64 \\
        weight decay & 0 \\
        \# stacks & 1 \\
        stack type & generic \\
        backcast loss ratio & 0.75 \\
        layer width & 512 \\
        MAML lr & 1e-1 \\
        learner lr (SN, BN, FR) & 1e-3, 1e-3, 1e-3 \\
        \bottomrule
    \end{tabular}
\end{table}

\subsubsection{Temporal Fusion Transformer}
\begin{table}[H]
    \centering
    \begin{tabular}{c|c}
        \toprule
        learning rate & 1e-5  \\
        batch size & 64 \\
        weight decay & 0 \\
        hidden size & 128 \\
        \# lstm layers & 2 \\
        \# attn heads & 4 \\
        hidden continuous size & 16 \\
        \# output quantiles & 3 (0.1, 0.5, 0.9) \\
        MAML lr & 1e-1 \\
        learner lr (SN, BN, FR) & 1e-4, 1e-3, 1e-3 \\
        \bottomrule
    \end{tabular}
\end{table}

\subsubsection{LSTM}
\begin{table}[H]
    \centering
    \begin{tabular}{c|c}
        \toprule
        learning rate & 1e-4  \\
        batch size & 64 \\
        weight decay & 1e-2 \\
        \# layers & 2 \\
        hidden state size & 10 \\
        MAML lr & 1e-1 \\
        learner lr (SN, BN, FR) & 1e-3, 1e-3, 1e-3 \\
        
        \bottomrule
    \end{tabular}
\end{table}

\subsubsection{MLP}
We use hidden layers of uniform size and ReLU activation
\begin{table}[H]
    \centering
    \begin{tabular}{c|c}
        \toprule
        learning rate & 1e-2  \\
        batch size & 64 \\
        weight decay & 0 \\
        \# hidden layers & 3 \\
        hidden size & 64 \\
        MAML lr & 1e-1 \\
        learner lr (SN, BN, FR) & 1e-2, 1e-4, 1e-3 \\
        
        \bottomrule
    \end{tabular}
\end{table}

\end{document}